\documentclass{iLRNConfProc}
\usepackage[backend=biber,style=apa,]{biblatex}
\usepackage{amsmath}
\DeclareLanguageMapping{english}{english-apa}
\title{El Niño Prediction Based on Weather Forecast \\and Geographical Time-series Data}

\author{
\begin{minipage}{\textwidth}
\centering
\textbf{Viet Trinh\textsuperscript{1,2,*}, Ha-Vy Luu\textsuperscript{1,2,†}, Quoc-Khiem Nguyen-Pham\textsuperscript{1,2,†}, \\Hung Tong\textsuperscript{1,2,†}, Thanh-Huyen Tran\textsuperscript{1,2,†}, and Hoai-Nam Nguyen Dang\textsuperscript{1,2,†}}\\
\textsuperscript{1}University of Economics and Law, Ho Chi Minh, Vietnam \\
\textsuperscript{2}Vietnam National University, Ho Chi Minh, Vietnam \\
\textit{* Corresponding: (Email: tqviet@uel.edu.vn)}\\
† These authors contributed equally to this work.\\
\end{minipage}
}

\date{}

\begin{document}
\maketitle
\thispagestyle{empty}

\iLRNAbstract{This paper proposes a novel framework for enhancing the prediction accuracy and lead time of El Niño events, crucial for mitigating their global climatic, economic, and societal impacts. Traditional prediction models often rely on oceanic and atmospheric indices, which may lack the granularity or dynamic interplay captured by comprehensive meteorological and geographical datasets. Our framework integrates real-time global weather forecast data with anomalies, subsurface ocean heat content, and atmospheric pressure across various temporal and spatial resolutions. Leveraging a hybrid deep learning architecture that combines a Convolutional Neural Network (CNN) for spatial feature extraction and a Long Short-Term Memory (LSTM) network for temporal dependency modeling, the framework aims to identify complex precursors and evolving patterns of El Niño events.}


\jelcodes{A14; M31; M37}

\section{INTRODUCTION} \label{sec-introduction}
The \textit{El Niño Southern Oscillation} (ENSO) phenomenon, or \textit{El Niño} for short, was formally defined by Trenberth in 1997 to represent a critical aspect of global climate variability \parencite{TheDefinitionofElNio}. Characterized by anomalous warming of sea surface temperatures (SST) in the eastern Pacific, particularly in the \textit{Niño 3.4 region}, El Niño events disrupt normal oceanic upwelling, leading to significant ecological and socioeconomic consequences. Historically, events such as the 1982–1983 El Niño have demonstrated profound impacts on marine ecosystems, notably affecting fish populations like the Peruvian anchovy, which once supported the world's largest fishery \parencite{doi:10.1126/science.222.4629.1203, NIQUEN2004563}. Beyond biological repercussions, El Niño also exerts considerable influence on economic systems, as evidenced by its effects on commodity markets such as Colombian coffee \parencite{https://doi.org/10.1111/agec.12447}. The severity and spatial extent of these impacts are most likely modulated by the geographical characteristics such as intensity and timing of oceanic anomalies \parencite{adgeo-6-221-2006}.

Accurate and timely prediction of El Niño events is thus crucial for mitigating adverse effects and informing effective response strategies. Traditional approaches to climate modeling, such as General Circulation Models (GCM), pioneered by Smagorinsky over four decades ago \parencite{GENERALCIRCULATIONEXPERIMENTSWITHTHEPRIMITIVEEQUATIONS}, have provided foundational insights into atmospheric and oceanic coupling. However, GCMs often struggle with adequately resolving fine-scale processes like cloud formation, moist convection mixing \parencite{wilby1997downscaling, Kochkov_2024, UnderstandingElNioinOceanAtmosphereGeneralCirculationModelsProgressandChallenges}. While Regional Climate Models (RCM) offer enhanced spatial resolution at reduced computational cost \parencite{LAPRISE20083641}, they are limited by one-way nesting methods, reliance on lateral boundary conditions, and the absence of two-way feedback mechanisms \parencite{2004}.

To address these limitations, this study proposes a novel dual deep learning framework for improved El Niño forecasting. By integrating a Convolutional Neural Network (CNN) and a Long Short-Term Memory (LSTM) architecture, this framework aims to leverage historical Sea Surface Temperature (SST) and Ocean Heat Content (OHC) data to identify complex precursors and evolving patterns indicative of El Niño onset and progression. This approach holds the potential to significantly enhance forecast accuracy and lead time, facilitating earlier warnings and better preparedness in case of irregular or El Niño event. The rest of this paper is organized as follows: Section \ref{sec-related} reviews relevant literature on El Niño forecasting. Section \ref{sec-method} details the methodology, encompassing data sources and the system's model architecture. Section \ref{sec-results} presents the experimental results and a comprehensive discussion. Finally, Section \ref{sec-conclusion} concludes the study and outlines directions for future research.

\section{RELATED WORKS} \label{sec-related}
The forecasting of El Niño Southern Oscillation events has progressed significantly since the pioneering work of Bill Quinn in the mid-1970s, whose method utilized the Southern Oscillation Index (SOI) to forecast a weak El Niño \parencite{doi:10.1126/science.191.4225.343}. In recent years, rapid technological advancements and the increasing availability of big data have facilitated the development of innovative and powerful machine learning (ML) models. These leverage extensive and diverse datasets to uncover complex patterns, improve predictive accuracy, and create more adaptive and intelligent systems across various domains, including climate science.

Deep learning approaches have shown considerable promise in time series analysis relevant to climate forecasting. For instance, \parencite{10.1145/3691338} provided a comprehensive summary of challenges associated with applying deep anomaly detection model to time series data. Subsequently, \parencite{doi:10.34133/olar.0012} introduced the Spatio-Temporal Information Extraction and Fusion (STIEF) model, an interpretable deep learning framework capable of forecasting the Niño 3.4 index with a lead time of 22–24 months. Concurrently, \parencite{srinivasan2020mlcnn} developed a Convolutional Neural Network model designed to identify anomalies in plots that closely resemble those detected by human reviewers, aiming to automate manual anomaly detection processes in quality control plots. 

More recently, hybrid architectures integrating Mamba and Transformer models have been applied to weather dynamics, addressing the challenges of long- and short-range time series forecasting \parencite{10823516}. This approach is crucial for accurately predicting future trends and patterns over extended periods, offering significant improvements in prediction accuracy, scalability, and memory efficiency. Furthermore, \parencite{https://doi.org/10.1029/2023GL105175} integrated a CNN model using cosine distance to predict zonal sea surface temperature anomaly (SSTA) patterns 12–16 months in advance, a method that played a critical role in determining model behaviors for El Niño heat map predictions.

Despite these advancements, several challenges persist in El Niño forecasting and its practical application. A key concern is the unclear relationship between forecast-based contingency planning and improved disaster preparedness and response, necessitating further research to understand how humanitarian organizations can effectively utilize these new types of information \parencite{TOZIERDELAPOTERIE201881}. Moreover, specific limitations within current ML-based El Niño prediction models have been identified, including issues with ENSO data frequency (e.g., monthly measurements compared to daily/weekly data), a lack of out-of-sample validation, and insufficient real-time forecast testing \parencite{https://doi.org/10.1002/for.2914}. Other examples include the impact of outliers in datasets due to unclearly processed models \parencite{10.3389/fphy.2019.00153}. While some proposed approaches have been effective in predicting El Niño events with a one-year lead time, they have struggled to accurately forecast extremely strong El Niño events \parencite{WANG2021104695}.

\section{METHODOLOGY}\label{sec-method}
\subsection{Data Collection and Pre-processing}
\subsubsection{Data Collection}
This study utilized two distinct oceanic datasets, \textbf{\textit{Sea Surface Temperature}} (SST) and \textbf{\textit{Ocean Heat Content}} (OHC), to train a deep learning system for identifying indicators of El Niño progression. The SST data were acquired from the National Oceanic and Atmospheric Administration’s (NOAA) Extended Reconstructed Sea Surface Temperature (ERSST) version 5. This dataset provides monthly global SST values, which are essential for comprehending ocean-atmosphere interactions, particularly within the context of El Niño dynamics. OHC data were obtained from the Oceanographic Data Center, Chinese Academy of Sciences, ensuring access to high-quality, standardized oceanographic measurements. Both SST and OHC datasets were provided in \textit{NetCDF} format, a widely adopted standard in climate and geospatial research, which facilitates efficient storage and access to multidimensional data. For the purpose of this study, the dataset was specifically curated to cover the period from January 2000 to September 2023. Furthermore, the geographical scope was precisely limited to the \textit{Niño 3.4 region} (5\textdegree S to 5\textdegree N latitude, 120\textdegree W to 170\textdegree W longitude). This region is central to the analysis of El Niño anomaly events; consequently, data outside these defined geospatial boundaries were excluded to maintain the model's relevance to the targeted area of investigation.

\subsubsection{Data Pre-processing}
\begin{figure}
	\centering
		\includegraphics[width=0.7\linewidth]{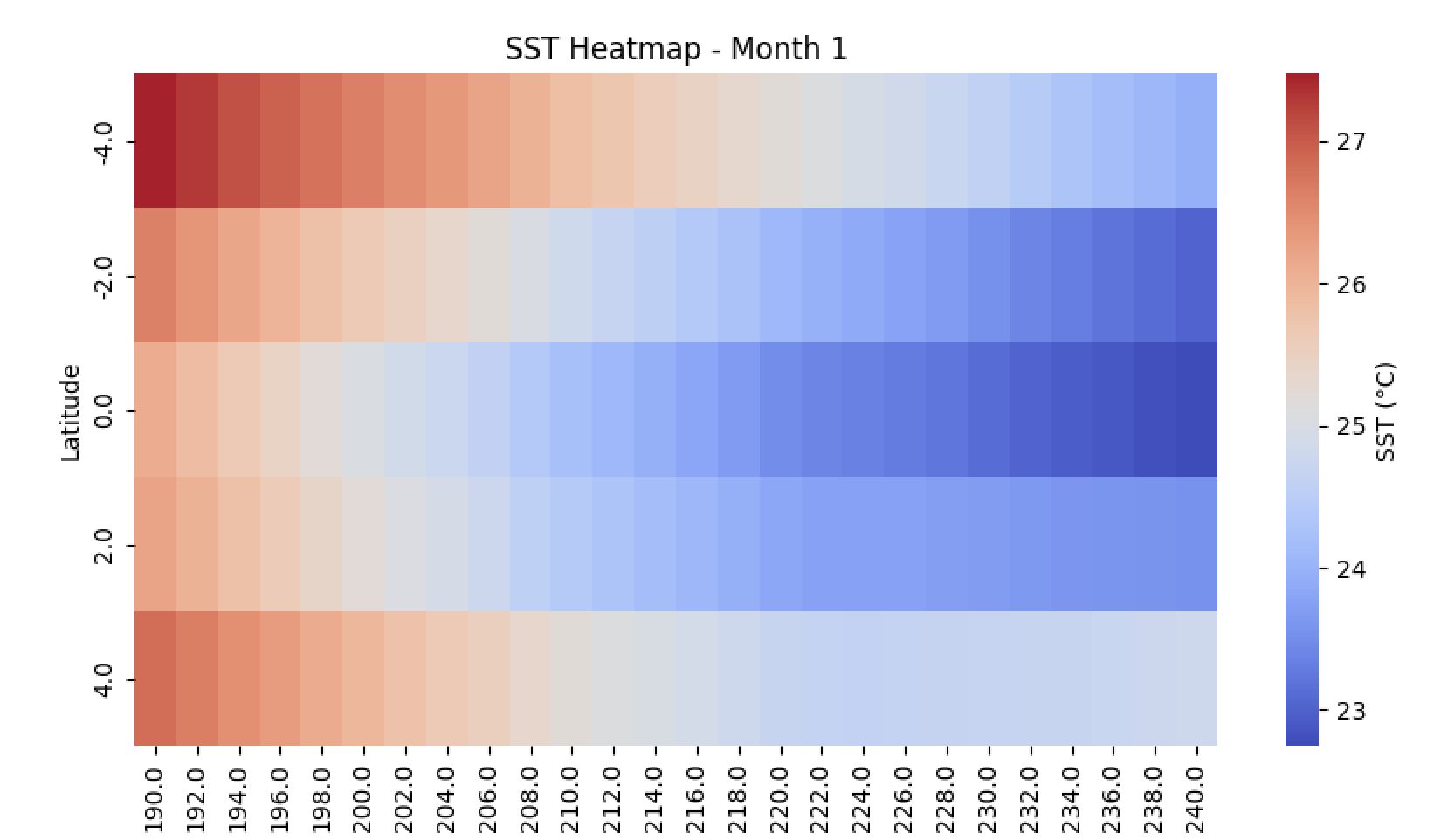}
	\caption{Heatmap represents a spatial distribution of SST anomalies in the Niño 3.4 region.}
	\label{fig:heatmap}
\end{figure}

Prior to ingestion into the proposed deep learning system for training and prediction, the raw SST and OHC data undergo a rigorous pre-processing pipeline, converting them into two supplementary formats: \textbf{\textit{heatmap}} and \textbf{\textit{normalized numeric-CSV}}. The \textit{heatmaps} serve as a visual representation of monthly SST values across the specified geographic grid, effectively encoding the spatial distribution of temperature anomalies. In these visualizations, warmer colors, such as reds and yellows, denote higher temperatures, while cooler hues, including blues and violets, indicate lower counterparts (Figure~\ref{fig:heatmap}). Subsequently, both SST and OHC data are \textit{normalized} using a Min-Max scaling mechanism to a range of $[0, 1]$. This normalization step is crucial for ensuring compatibility in data range and distribution, facilitating seamless integration, and preventing features with larger magnitudes from disproportionately influencing the model's learning process. 

These generated heatmaps are indispensable for the CNN component of the system, enabling it to learn intricate spatial patterns over time. This capability allows the CNN to effectively detect anomalies and variations in SST values across both latitudinal and longitudinal dimensions. Conversely, the normalized SST and OHC datasets are utilized by the LSTM network to model temporal patterns, thereby capturing the sequential dependencies within the oceanic data.

\subsection{Oceanic Niño Index}
The Oceanic Niño Index (ONI) serves as a pivotal metric for identifying the occurrence of El Niño events. This index is derived from SST anomalies within the crucial Niño 3.4 region, specifically when these anomalies meet or exceed a threshold of $0.5^{\circ}\text{C}$. The ONI is calculated as a 3-month running mean of these regional SST anomalies, providing a smoothed representation that helps to distinguish sustained El Niño conditions from transient fluctuations. 

Given a particular time $t$, a geographical coordinates ($lat$, $lng$), and a number of $N$ data points, a regional SST anomaly $A_{region}(t)$ can be determined as:
\[
A_{region}(t) = \frac{1}{N} \sum_{(lat,lng)\ \in \ \text{Niño 3.4}} T_{lat,lng}(t) - \frac{1}{30} \sum_{y_t=c(y_t)-15}^{c(y_t)+14} T_{lat,lng}(y_t,m(t)) 
\]

Where:
\begin{itemize}
  \item $T_{lat,lng}(t)$ is the observed SST at coordinates ($lat$, $lng$) at time $t$
  \item $T_{lat,lng}(y_t,m(t))$ is the average SST for month $m(t)$ of the center year $c(y_t)$, corresponding to the year of time $t$ at coordinates ($lat$, $lng$)
\end{itemize}

The second term of the equation refers to \textit{climatology}, which represents the 30-year patterns and variations of average SST within a specific region. Thus, the ONI value at a particular time $t$ can be calculated as:
\[\text{ONI}(t) = \frac{1}{3}\left( A_{region}(t-2) + A_{region}(t-1) + A_{region}(t)\right)\]

\subsection{System Pipeline}
\begin{figure}[htbp]
    \centering
    \includegraphics[width=\linewidth]{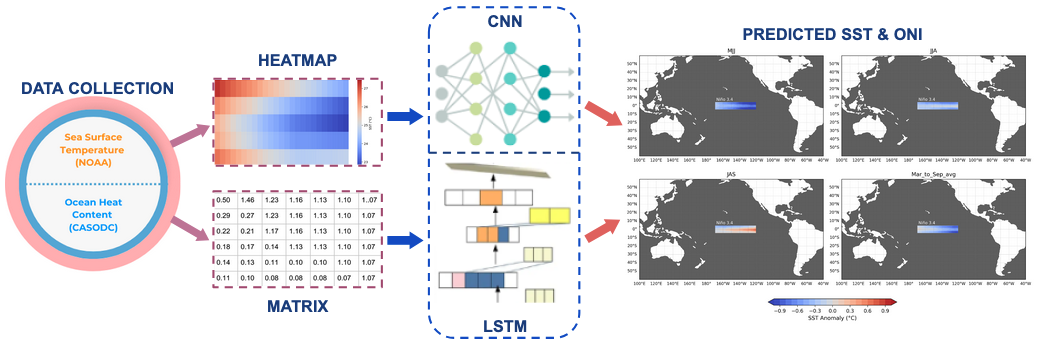}
    \caption{The proposed framework collects historical Sea Surface Temperature (SST) and Ocean Heat Content (OHC) data values; converts the oceanic data into heatmap and numeric-CSV representations; leverages CNN-based and LSTM-based architectures to extract spatial and temporal patterns; and predicts future SST to identify El Niño events in the Niño 3.4 region based on the calculated Oceanic Niño Index (ONI).}
    \label{fig:pipeline}
\end{figure}

This research investigates the efficacy of deep learning techniques for forecasting El Niño events within the Niño 3.4 region. The proposed system integrates both spatial and temporal modeling of oceanographic data, leveraging the inherent strengths of CNN and LSTM architectures. Figure \ref{fig:pipeline} provides a detailed illustration of the system's pipeline. Within this framework, CNN is employed to extract spatial patterns from SST heatmaps, thereby capturing temperature anomalies across specific geographic grids. Conversely, LSTM is utilized to model the temporal dynamics inherent in both SST and OHC time series, enabling the system to track evolving oceanic conditions over time. The outputs from both network architectures are then synthesized to assess the likelihood of an impending El Niño event. This synergistic integration of CNN and LSTM allows the framework to capitalize on their complementary strengths, effectively mitigating potential biases that might arise from relying on a single network type.

\subsubsection{ConvLSTM-XT Architecture}
\begin{figure}[htbp]
	\centering
	\includegraphics[width=\linewidth]{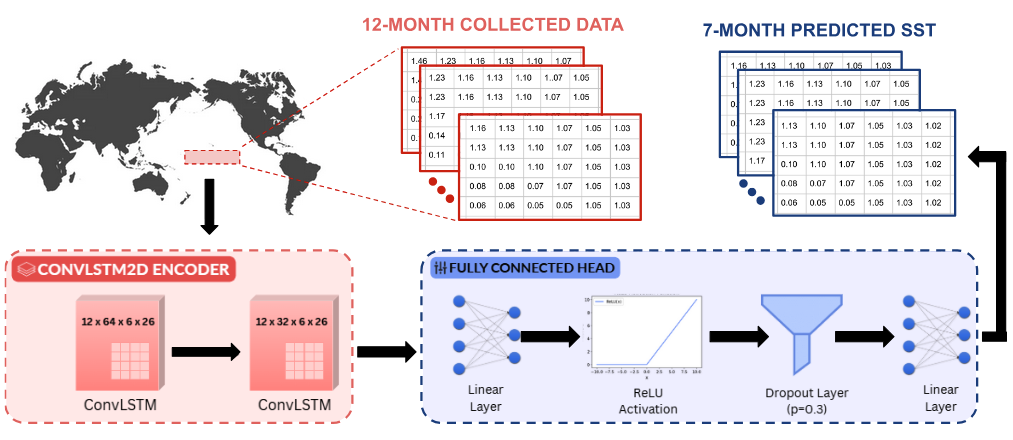}
	\caption{The ConvLSTM-XT model is designed for high-accuracy spatio-temporal forecasting of SST using historical gridded observational data.}
	\label{fig:architecture}
\end{figure}

The Long Short-Term Memory (LSTM) architecture employed in this study comprises two ConvLSTM blocks and a single fully-connected head, designed to enhance the comprehensive understanding of the ocean's thermal state through both SST and OHC input streams (Figure~\ref{fig:architecture}). Each ConvLSTM block extends the conventional LSTM cell by substituting matrix multiplications with 2D convolutions. This modification enables the block to process spatial information while simultaneously tracking temporal evolution, a crucial feature for analyzing dynamic oceanographic data. The final fully-connected layer incorporates a Rectified Linear Unit (ReLU) activation function and a dropout rate of 0.3. This layer flattens the output into a 4D tensor, which represents the predicted SST grids for each time step within the defined forecast horizon.

\section{RESULTS AND DISCUSSION}\label{sec-results}
\subsection{Training and Evaluation}
The ConvLSTM-XT model underwent training for 50 epochs, utilizing the Adam optimizer with an initial learning rate of 0.001 and a batch size of 32. Mean Squared Error (MSE) was employed as the loss function to optimize the model's predictive accuracy for SST values. The dataset was partitioned into an 80\% training set and a 20\% testing set.

Model performance was rigorously benchmarked against the Oceanic Niño Index (ONI) threshold for five consecutive overlapping quarters, encompassing 52 monthly observations from November 2018 to September 2023. For both the Convolutional Neural Network (CNN) and Long Short-Term Memory (LSTM) components, a 52×5 matrix was constructed. This matrix represents 52 time steps across five overlapping quarters, each spanning seven months due to a sliding window approach. The final forecasted anomaly for evaluation was derived from the ensemble average of the quarterly anomalies from both models. This integrated approach leverages the complementary strengths of the CNN, which captures surface temperature dynamics, and the LSTM, which incorporates subsurface ocean heat content, thereby enhancing the overall robustness of the prediction system.

\subsubsection{Evaluation Configurations}
Six configurations are defined to assess performance across varying forecast horizons:
\begin{itemize}
    \item \textit{Configuration 0}: All five quarters use observed anomalies (baseline, no forecasting)
    \item \textit{Configurations 1-4}: A mix of observed (5-$k$) quarters and forecasted ($k$) quarters, where $k$ ranges from 1 to 4
    \item \textit{Configuration 5}: All five quarters use the averaged forecasted anomalies
\end{itemize}

For each configuration, predictions are compared against observed El Niño occurrences across the 52 time steps. A confusion matrix is constructed with the following components:
\begin{itemize}
    \item \textit{True Positive (TP):} Correct predictions where both forecasted and observed anomalies exceed 0.5\textdegree C for all five quarters
    \item \textit{True Negative (TN):} Correct predictions where both forecasted and observed anomalies are below 0.5\textdegree C for at least one quarter
    \item \textit{False Positive (FP):} Incorrect predictions where forecasted anomalies exceed 0.5\textdegree C, but observed anomalies do not
    \item \textit{False Negative (FN):} Incorrect predictions where forecasted anomalies are below 0.5\textdegree C, despite observed anomalies exceeding the threshold
\end{itemize}

Accuracy of each prediction is then calculated as:
\[ \text{Accuracy} = \frac{\text{TP} + \text{TN}}{\text{TP} + \text{TN} + \text{FP} + \text{FN}}
\]

\subsection{Results and Discussion}
\begin{figure}[htbp]
    \centering
    \includegraphics[width=\linewidth]{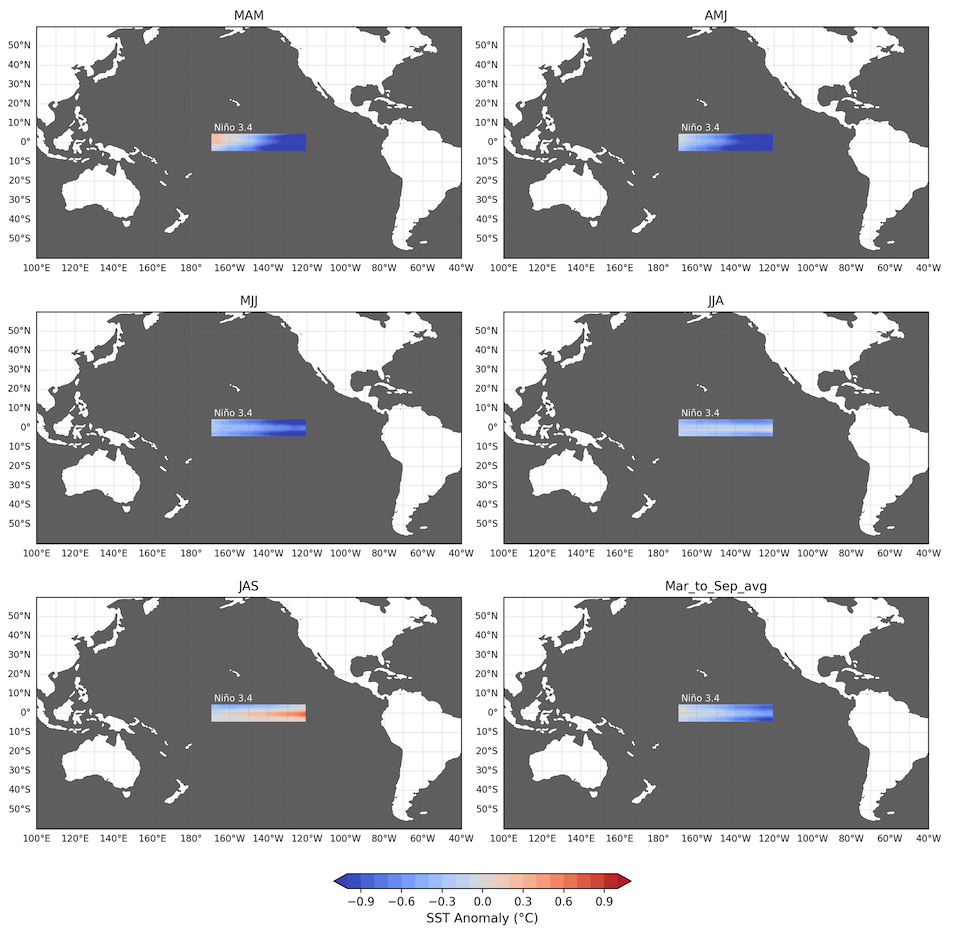}
    \caption{Qualitative analysis of predicted SST anomaly heatmaps from March to September 2023.}
    \label{fig:results}
\end{figure}

Figure~\ref{fig:results} presents a visual assessment of the model's spatio-temporal forecasting skill for SST anomalies within the critical Niño 3.4 region during the March to September 2023 period. The heatmaps depict the temporal evolution of predicted SST anomalies across sequential three-month overlapping periods (March-May [MAM], April-June [AMJ], May-July [MJJ], June-August [JJA], July-September [JAS]), which was a pivotal transition year for El Niño development.

\begin{itemize}
    \item \textit{Initial Phases (MAM, AMJ, MJJ)}: During these earlier forecast periods in 2023, the Niño 3.4 region predominantly exhibits cold (blue) anomalies, indicating the model's prediction of cooler-than-average SSTs. This suggests the model effectively captured the lingering cold conditions that preceded the 2023 El Niño onset.
    \item \textit{Transitional Phases (JJA, JAS)}: As the forecast progresses into the JJA and JAS periods of 2023, a discernible shift toward warmer (red/orange) anomalies begins to emerge within the Niño 3.4 region. While these anomalies may not consistently or strongly exceed the 0.5\textdegree C El Niño threshold, their appearance signifies the model's prediction of a warming trend, qualitatively aligning with the observed development of the 2023 El Niño.
    \item \textit{Average Trend (Mar\_to\_Sep\_avg)}: The aggregated average from March to September 2023 further accentuates this overall warming trend in the equatorial Pacific, though the average signal may be attenuated by the colder anomalies predicted during the earlier months of the period.
\end{itemize}

This visual sequence qualitatively suggests the model's capability to predict a developing warming trend in the central Pacific, which is characteristic of El Niño onset. Despite the evident warming signal in the heatmaps, the stringent criterion requiring all five consecutive three-month periods to exceed 0.5\textdegree C likely results in the model consistently classifying borderline or nascent warming events as non-El Niño.

\begin{table}[htbp]
    \centering
    \caption{El Ni\~{n}o prediction's accuracy across different forecast configurations}
    \label{tab:accuracy}
    \begin{tabular}{clc}
        \toprule
        \textbf{No.} & \textbf{Forecast configuration} & \textbf{Accuracy (\%)} \\
        \midrule
        1 & 4 observed quarters + 1 forecasted quarter & 90.57 \\
        2 & 3 observed quarters + 2 forecasted quarters & 90.57 \\
        3 & 2 observed quarters + 3 forecasted quarters & 90.57 \\
        4 & 1 observed quarter + 4 forecasted quarters & 90.57 \\
        5 & 0 observed quarter + 5 forecasted quarters & 83.02 \\
        \bottomrule
    \end{tabular}
\end{table}

Table \ref{tab:accuracy} details the proposed system's accuracy in forecasting El Niño events across various forecast horizons, categorized by the interplay between observed and predicted quarters. The model consistently demonstrates an accuracy of 90.57\% for configurations 1 through 4, where the prediction integrates a combination of observed and forecasted data. This sustained high accuracy suggests robust performance even as the proportion of forecasted quarters increases. However, a noticeable decline in accuracy to 83.02\% is observed in configuration 5, where all five quarters are entirely predicted without the inclusion of observed data. This reduction indicates a limitation in the model's performance when it relies solely on its own extrapolations.

The consistent 90.57\% accuracy across configurations 1 to 4 underscores the ConvLSTM-XT architecture's effectiveness in integrating observed data with short to intermediate-term predictions. This robustness is likely attributable to the model's capacity to leverage spatial patterns via the CNN component and temporal dependencies via the LSTM network, particularly when anchored by real-world observations. Conversely, the accuracy drop to 83.02\% in configuration 5 highlights a limitation in the model's forecasting capability over longer horizons without direct observational input. This suggests that the model may struggle to fully capture the complex, long-term dynamics of El Niño events when operating exclusively on predicted data. 

Potential contributing factors to this decline include limitations in the input feature set, and architectural constraints. Specifically, the current reliance solely on Sea Surface Temperature and Ocean Heat Content may not fully encompass the broader oceanic and atmospheric interactions that drive El Niño phenomena. Extending the input feature set beyond SST and OHC to incorporate other relevant atmospheric and oceanic variables would allow the model to account for additional factors influencing SST. Variables such as surface wind stress, sea level anomaly, ocean currents, and bathymetry could provide a richer context, capturing more complex ocean-atmosphere interactions and improving model robustness. Additionally, the proposed model's architecture might have inherent limitations in accurately extrapolating multi-quarter predictions without the stabilizing influence of observed data. Further research could explore the integration of additional climate variables or advanced architectural modifications to enhance long-term predictive accuracy.

\section{CONCLUSION} \label{sec-conclusion}
In conclusion, this research presents an integrated deep learning approach designed to enhance the anticipation of El Niño occurrences through the sophisticated modeling of both spatial and temporal climate patterns. The proposed ConvLSTM-XT architecture, trained on historical Sea Surface Temperature and Ocean Heat Content data, demonstrates robust predictive accuracy, even across extended forecast windows. While these results are promising, particularly in scenarios that blend real-world observations with predicted data, challenges persist when the model operates without observational anchors. Future improvements will critically depend on enhancing data diversity and refining model granularity.

The potential applications of this model extend significantly beyond academic research, offering profound global implications. This framework could be adapted to serve as an early warning system for predicting El Niño events, thereby providing governments and organizations with crucial advanced notice of potential climatic disruptions. Furthermore, the model could prove instrumental in policy planning for climate adaptation and mitigation strategies, empowering regions to prepare for the multifaceted economic and environmental impacts associated with such events.

\printbibliography

\end{document}